\documentclass[preprint]{elsarticle}

\usepackage{amssymb}
\usepackage{amsthm}
\usepackage{amsmath}
\usepackage{bm}

\newtheorem{thm}{Theorem} 
\newtheorem{lem}[thm]{Lemma}
\newdefinition{definition}{Definition} 
\newproof{prf}{Proof}
\newproof{pot}{Proof of Theorem \ref{thm2}}

\journal{Information and Computation}

\begin{document}

\begin{frontmatter}



\title{On the equivalence of Occam algorithms}


\author[inst1]{Zaman Keinath-Esmail}
\affiliation[inst1]{organization={Rudolf Peierls Centre for Theoretical Physics, University of Oxford},
            addressline={Beecroft Building Parks Road}, 
            city={Oxford},
            postcode={OX1 3PU},
            country={UK}}

\begin{abstract}
Blumer et al. (1987, 1989) showed that any concept class that is learnable by Occam algorithms is PAC learnable. Board and Pitt (1990) showed a partial converse of this theorem: for concept classes that are closed under exception lists, any class that is PAC learnable is learnable by an Occam algorithm. However, their Occam algorithm outputs a hypothesis whose complexity is $\delta$-dependent, which is an important limitation. In this paper, we show that their partial converse applies to Occam algorithms with $\delta$-independent complexities as well. Thus, we provide \textit{a posteriori} justification of various theoretical results and algorithm design methods which use the partial converse as a basis for their work.
\end{abstract}

\begin{keyword}
Occam algorithms \sep polynomial learnability \sep PAC learning \sep statistical learning theory
\MSC[2010] 68Q32 \sep 68W01

\ 

\textit{Funding:} This research did not receive any specific grant from funding agencies in the public, commercial, or not-for-profit sectors.

\end{keyword}

\end{frontmatter}


\section{Introduction}
\label{sec:introduction}

For decades, there has been debate regarding the necessity of simplicity for machine learning \cite{pearl1978connection,domingos1999role,webb1996further}. Many of these analyses have focused on the implications and uses of complexity-based algorithms defined by Blumer et al. in two seminal papers \cite{blumer1987occam,blumer1989learnability}. Their algorithms were defined such that they achieved zero training error on a sample, and outputted a hypothesis whose complexity (VC dimension for continuous alphabets; description length for discrete ones) was at most a polynomial in the target concept complexity, multiplied by a sublinear factor in the sam. These ``Occam algorithms" are weak approximations of the minimum-consistent-hypothesis problem \cite{board1990necessity}. In this paper, we focus on the continuous-alphabet Occam algorithms.

In 1989, Blumer et al. \cite{blumer1989learnability} showed that if a concept was learnable by their Occam algorithm, then it was polynomially learnable; they left open the question of whether the converse of this theorem was true. Board and Pitt \cite{board1990necessity} proved a partial converse, namely that, for concept classes closed under exception lists, if a class was polynomially learnable in general, then it was learnable by an Occam algorithm. This proof suggested that for concept classes which are closed under exception lists, learnability is equivalent to weak approximability of the minimum-consistent hypothesis problem \cite{board1990necessity}. This equivalence formed the basis for subsequent theoretical work analysing the learnability of DFAs \cite{pitt1993minimum} and decision trees and lists \cite{hancock2005lower}, as well as motivating the design of practical algorithms such as prediction algorithms for optimal data prefetching \cite{vitter1996optimal}.

However, the Occam algorithm model that was used by Board and Pitt \cite{board1990necessity} was different to the one defined by Blumer et al. \cite{blumer1989learnability}: the former is a construction based on a functional algorithm requiring $\epsilon$, $\delta$, $s$, and the training sample as explicit inputs, while Blumer et al.'s model requires only the training sample \cite{blumer1989learnability}. While it has been shown that many frequently studied PAC algorithms form an equivalence class \cite{haussler1991equivalence}, it is not clear whether Board and Pitt's \cite{board1990necessity} version is a member of this class. Further, Board and Pitt later remove a degree of freedom by defining $\epsilon$ in terms of $m$. In some cases, changing degrees of freedom can alter whether an algorithm always halts or just usually halts: even if Board and Pitt's model of learnability falls under the aforementioned equivalence class, it is unclear whether the equivalent algorithm in Blumer et al.'s definition would always halt \cite{haussler1991equivalence,board1990necessity,blumer1989learnability}.

In addition, the complexity of the output of Board and Pitt's Occam algorithm is $\delta$-dependent \cite{board1990necessity}, while in Blumer et al.'s definition it is not \cite{blumer1989learnability}. Thus, it is not immediately clear whether Board and Pitt's \cite{board1990necessity} conclusions hold for more commonly used polynomial PAC learning models and $\delta$-independent output complexities (and are hence a general property of polynomial learnability and Occam algorithms), or whether their theorem results purely from the specific choice of algorithmic model. Therefore, it may not be justified to use their results as the basis for further research on general learnability and algorithm construction, as is currently the case \cite{vitter1996optimal,hancock2005lower,pitt1993minimum}. 

In this paper, we show that the polynomial PAC model used by Board and Pitt \cite{board1990necessity} is equivalent to the models used by Blumer et al. and Haussler et al. \cite{blumer1989learnability,haussler1991equivalence}, and that we can construct an Occam algorithm from such a functional algorithm so that its output complexity is $\delta$-independent, thereby explicitly linking the work of Board and Pitt \cite{board1990necessity} to that of Blumer et al. \cite{blumer1989learnability}. Our theorem provides the link needed to justify the use of Board and Pitt's \cite{board1990necessity} work as a theoretical basis from which to further analyse questions of learnability and algorithm design by showing that their result holds irrespective of the specific algorithmic model they chose.

We will begin by defining concept classes, parameters such as $s$ and $\delta$, and our choice of complexity measure, the Vapnik-Chervonenkis (VC) dimension, along with relevant lemmas. Next, we define functional and oracle algorithmic models and provide an equivalence theorem due to Haussler et al. \cite{haussler1991equivalence} linking the two. We proceed to provide the definitions of Occam algorithms used by Blumer et al. \cite{blumer1989learnability} and Board and Pitt \cite{board1990necessity}, pointing out the differences and showing that the latter is not obviously equivalent to the former. Finally, we prove that the two Occam algorithm definitions are, in fact, equivalent, thus justifying general analyses of learnability based on Board and Pitt's work \cite{board1990necessity}.

\subsection{A note on algorithm names}
\label{sec:note-on-names}

It is worth pausing before the next section to explicitly list the algorithmic models we will be discussing in the remainder of this paper. We will distinguish between four models: $\delta,s$-known functional algorithms, $\delta,s$-unknown functional algorithms, $\delta$-dependent Occam algorithms, and $\delta$-independent Occam algorithms.

The $\delta,s$-unknown functional algorithm is the standard definition of a functional algorithm used by Blumer et al. and Haussler et al. \cite{blumer1989learnability,haussler1991equivalence}, given in Definition \ref{def:functional_poly-learnability}. The $\delta,s$-known functional algorithm is a version used only by Board and Pitt \cite{board1990necessity}. We will need to show an equivalence between these two models, as the $\delta,s$-known functional algorithm is not one of the models contained in the equivalence class provided by Haussler et al. in Theorem \ref{theorem:equivalence-polynomial}. This result will be Lemma \ref{lemma:dependent-independent}.

The $\delta$-dependent Occam algorithms and $\delta$-independent Occam algorithms, on the other hand, refer not to whether $\delta$ is an explicit input, but whether the VC dimension of the hypothesis space is polynomially dependent on $\frac{1}{\delta}$ or not. The $\delta$-independent version is used by Blumer et al. \cite{blumer1989learnability} while Board and Pitt \cite{board1990necessity} use a $\delta$-dependent version. The main result of our paper will be extending the analysis conducted by Board and Pitt to apply to $\delta$-independent Occam algorithms as well, resulting in Theorem \ref{thm:main}.

\section{Notation and definitions}
\label{sec:notation-definitions}

\subsection{Terminology}
\label{sec:concept-classes}

Let $X$ be a stratified learning domain (i.e. we can write it as $X=\cup_{n\geq1}X_{n}$ where each $X_{n}$ is a region in $n$-dimensional Euclidean space). A \textit{concept} is a subset of $X$; we can consider it as labelling all elements of $X$ with either a 1 or a 0, depending on whether the element is in the concept or not.

Then let $C\subseteq2^{X}$ be a well-behaved\footnote{in a measure-theoretic sense made explicit in Blumer et al's work \cite{blumer1989learnability}.} class of concepts on $X$ and let the \textit{concept complexity measure} $\textbf{size}: C\rightarrow\mathbb{Z^{+}}$. Let there also be associated with $C$ a set of representations for concepts in $C$ given in some representation language $\Gamma$. We assume there is a function $\sigma$ from the set of representations in $\Gamma$ onto $C$, that $\sigma$ is in $\mathbf{P}$ (or in $\mathbf{RP}$), and that given $x \in X$ and a string in $\Gamma$, we can decide in polynomial time if $x$ is in the concept represented by the string (i.e. $C$ is polynomially evaluable). Finally, we say that $C$ is \textit{stratified}, i.e. $C=\cup_{n\geq1}C_{n}$, where each $C_{n}$ is a region in $n$-dimensional Euclidean space.

By $\mathcal{C}$ we denote $\left(X,C\right)$ along with their function $\textbf{size}$, $\sigma$ and $\Gamma$, so $\mathcal{C}=\left(X,C,\Gamma,\sigma,\textbf{size}\right)$. We include $\Gamma$ because learnability may in fact be a property of the representation of a set of concepts, not just the concepts themselves \cite{board1990necessity}. We use the term \textit{concept class} to refer to $\mathcal{C}$ as well as to $C$. We say a concept $c$ is in $\mathcal{C}$ if $c \in C$. 

We define a hypothesis class $\mathcal{H}$ where, as with $\mathcal{C}$, $\mathcal{H} = \left(X,H,\Gamma,\sigma,\textbf{size}\right)$, where $H \subseteq2^{X}$ is also a class of concepts on $X$ but need not be related to $C$. 

We use $\mathcal{P}$ to represent probability distributions on the domain $X$. A sample set taken according to $\mathcal{P}$ on $X$ is denoted $M$ and contains $m$ tuples (or points) $(x,c(x))$; we say $M$ has cardinality $m$. Denote the set of $x$-values in $M$ by the set $Z$. The error, or accuracy parameter, $\epsilon$ of a hypothesis $h$ with respect to a concept $c$ is $\epsilon=\mathcal{P}(h\triangle c)$ where $\triangle$ denotes symmetric difference. In words, $\epsilon$ is the probability due to $\mathcal{P}$ of the symmetric difference of the hypothesis and the concept. 

We cannot directly measure $\epsilon$ as it is a property over the entire domain, and the concept $c$ is unknown. Therefore, it is useful to define a training error $\epsilon'=\ell(h(Z)\triangle c(Z))$ for some loss function $\ell$. We say a hypothesis is consistent if $\epsilon'$ is zero on a sample. Then $\delta$ is the probability that the algorithm fails to return a consistent hypothesis.

Finally, we assume that $\mathbf{RP}\neq\mathbf{NP}$; otherwise the learnability of any concept class would be trivial.

\subsection{Vapnik-Chervonenkis dimension}
\label{sec:vc-dimension}

We have in Section \ref{sec:concept-classes} roughly stated the existence of a concept complexity measure, \textbf{size}, but we have not described any of its properties other than its domain and range. For the arguments presented in this paper we do not require any specific concept complexity measure function for \textbf{size}; it can be semantic (measuring breadth of explanatory power) or syntactic (e.g. the length of the description of a hypothesis in $\Gamma$), or something else entirely. We do make use of the (assumed) property that an algorithm cannot output a hypothesis with \textbf{size} greater than its runtime \cite{board1990necessity}. What will primarily matter, though, is how we can map from \textbf{size} to a combinatorial semantic complexity measure called the Vapnik-Chervonenkis (VC) Dimension \cite{board1990necessity,blumer1989learnability}.

\begin{definition}
\label{def:VCdim}
Given a nonempty concept class $\mathbf{C}=(C,X)$ with $C\subseteq2^{X}$ and a set of points $S\subseteq X$, $\Pi_{C}(S)$ denotes the set of all subsets of $S$ that can be obtained by intersecting $S$ with a concept in $C$. That is, $\Pi_{C}(S)=\{S\cap c:c\in C\}$.
If $\Pi_{C}(S)=2^{S}$ we say that $S$ is \textit{shattered} by $C$. The \textit{VC Dimension} of $C$ is the cardinality of the largest set of points $S\subseteq X$ that is shattered by $C$. If the set of shattered points is arbitrarily large, then the VC Dimension of $C$ is infinite.
$C$ is trivial if it consists of only one concept, or two disjoint concepts $c_{1}$ and $c_{2}$ such that $c_{1}\cup c_{2}=X$ (such concept classes require only one example to learn) \cite{blumer1989learnability}.
\end{definition}

\begin{lem}
For a finite hypothesis class $\mathcal{H}$, the VC dimension of $\mathcal{H}$ is bounded by $\text{VCdim}(\mathcal{H})\leq\log(|H|)$ \cite{understanding-ml}.
\end{lem}

\begin{lem}
\label{lemma:sauer-shelah-perles}
Let $\mathcal{H}$ be a hypothesis class with $\text{VCdim}(\mathcal{H})\leq d\leq\infty)$ and $\tau_{\mathcal{H}}(m)=\text{max}|\mathcal{H}_{M}|=|\Pi_{H}(M)|$ be the size of the effective hypothesis space; namely, the number of functions from a training set of size $M$ to $\{0,1\}$ that can be obtained by restricting $\mathcal{H}$ to $M$. Then $\forall\ m$, $\tau_{\mathcal{H}}(m)\leq\Sigma_{i=0}^{d}\binom{m}{i}$, and for $m\geq d+1$ we have $\tau_{\mathcal{H}}(m)\leq(em/d)^{d}$ \cite{understanding-ml}. 
Partially restated, we can say that $\tau_{\mathcal{H}}(m)\leq|S|^{d}+1$ \cite{board1990necessity}.
\end{lem}

\begin{lem}
\label{lemma:probability-of-inconsistency}
Let $H\subseteq2^{X}$ be a nonempty hypothesis class, $\mathcal{P}$ be a probability distribution on $X$, and target concept $c\subseteq X$. Then for any $\epsilon>1$ and $m\geq1$, given $m$ independent examples drawn according to $\mathcal{P}$, the probability that $\exists$ a hypothesis in $H$ that is consistent with all of these examples and has error greater than $\epsilon$ is at most $2\tau_{\mathcal{H}}(2m)2^{-\epsilon m/2}$ \cite{blumer1989learnability}.
\end{lem}

\begin{lem}
\label{lemma:exception-list-dimension}
Let $\mathbf{H}=(H,X)$ have VC dimension $d$. Let $H^{\triangle,l}=\{h\triangle E:h\in H,E\subseteq X,|E|\leq l\}$. If $d_{l}\geq2$ is the VC dimension of $H^{\triangle,l}$, then $\frac{d_{l}}{\log(d_{l})}\leq d+l+2$ \cite{board1990necessity}.
\end{lem}

\subsection{Functional and oracle algorithms}
\label{sec:functional-oracle}

Definitions similar to the following can be found in papers by Blumer et al., Valiant, and others \cite{blumer1989learnability,haussler1991equivalence,pitt1988computational}.

\begin{definition}[Functional polynomial learnability]
\label{def:functional_poly-learnability}
For a concept class $\mathcal{C}$ defined as in \ref{sec:concept-classes}, we say that $\mathcal{C}$ is \textit{properly polynomially learnable (poly-learnable) in the functional model} if $\exists$ polynomial-time learning algorithm $A$ that takes as input a sample of a concept in $\mathcal{C}$, outputs a hypothesis in $\mathcal{C}$, and has the property that, $\forall\ 0<\epsilon,\delta<1$ and $n,s\geq1,\exists$ sample size $m\left(\epsilon,\delta,n,s\right)$, polynomial in $1/\epsilon,1/\delta,n,\text{and}\ s$, such that, $\forall$ target concepts $c\in C_{n}$ with $\textbf{size}(c)\leq s$ and $\forall$ probability distributions $P$ on $X$, given a random sample of $c$ of size $m\left(\epsilon,\delta,n,s\right)$ drawn independently according to $P$, $A$ produces, with probability at least $1-\delta$, a hypothesis $h\in C_{n}$ that has error at most $\epsilon$ \cite{blumer1989learnability}.
\end{definition}

Functional polynomial learning algorithms receive only the sample of size $m$; it has no access to $\epsilon,\delta,$ or $s$ \cite{haussler1991equivalence}. It can always determine $n$ from the dimension of the data points it is given \cite{blumer1989learnability}. However, in the oracle model, the algorithm receives at least $\epsilon$ and $\delta$ as input and has access to an oracle $EX()$ that with each call returns a point in $X$ along with a label $0$ or $1$ indicating whether or not the point is in a fixed target concept $c\in C_{n}$ with $\textbf{size}\left(c\right)\leq s$. Each oracle call takes unit time. If $A$ is probabilistic, each call to a fair coin flip also takes unit time. After some time, $A$ halts and outputs a hypothesis in $C$ \cite{blumer1989learnability, haussler1991equivalence}.

\begin{definition}[Oracle polynomial learnability]
\label{def:oracle_poly-learnability}
For a concept class $\mathcal{C}$ defined as in \ref{sec:concept-classes}, we say that $\mathcal{C}$ is \textit{properly polynomially learnable (poly-learnable) in the oracle model} if $\exists$ algorithm $A$ that has the property that, $\forall\ 0<\epsilon,\delta<1$ and $n,s\geq 1,\exists$ time bound $T_{A}\left(\epsilon,\delta,n,s\right)$, polynomial in $1/\epsilon,1/\delta,n,\text{and}\ s$, such that, $\forall$ target concepts $c\in C_{n}$ with $\textbf{size}(c)\leq s$ and $\forall$ probability distributions $P$ on $X$, $A$ runs in time $T_{A}\left(\epsilon,\delta,n,s\right)$ and produces, with probability at least $1-\delta$, a hypothesis $h\in C_{n}$ that has error at most $\epsilon$ \cite{blumer1989learnability}.
\end{definition}

These models of learnability can easily be extended to situations in which we are learning $\mathcal{C}$ by $\mathcal{H}$ as opposed to just $\mathcal{C}$ by itself \cite{blumer1989learnability,haussler1991equivalence}. The following theorem, due to Haussler et al. \cite{haussler1991equivalence}, shows the equivalence between the different models of polynomial learnability. The property ``usually halts" refers to probabilistic oracle algorithms that use fair coin tosses to decide whether to call the oracle again or halt.

\begin{thm}\label{theorem:equivalence-polynomial}
    If $\mathcal{C}$ is learnable by hypothesis class $\mathcal{H}$ in any of the following algorithmic models, it is learnable by $\mathcal{H}$ in all of them:
    \begin{enumerate}
        \item \textbf{functional}($p_1$)
        \item \textbf{oracle}($p_1,p_2,p_3$),
    \end{enumerate}
    where we have properties:
    \begin{itemize}
        \item $p_1\in\{\textit{deterministic},\textit{randomised}\}$
        \item $p_2\in\{\textit{s-known},\textit{s-unknown}\}$
        \item $p_3\in\{\textit{always halts},\textit{usually halts}\}$
    \end{itemize}
    and the stipulation that $p_2=$ ``s-unknown" only when $p_3=$ ``usually halts" \cite{haussler1991equivalence}.
\end{thm}

\begin{definition}[Polynomial PAC learnability]
\label{def:poly-learnable}
    If a class is learnable by one (and hence all) of the algorithmic models in Theorem \ref{theorem:equivalence-polynomial}, then we simply call it ``polynomially PAC learnable" without specifying the specific model in which it is learnable.

\end{definition}

\subsection{Occam algorithms}
\label{sec:def-occam-algorithms}

The following definition is used by Blumer et al. in their paper introducing VC-dimension-based Occam algorithms \cite{blumer1989learnability}. Note that this is clearly a functional algorithm, matching definition \ref{def:functional_poly-learnability}.

\begin{definition}[$\delta$-independent Occam algorithm]
\label{def:s-delta-independent-occam}
    Let $A$ be a polynomial learning algorithm that, given a sample of a concept $c\in\mathcal{C}$, produces a hypothesis consistent with the sample. For concept class $\mathcal{C}$ and all $s,n,m\geq1$, let $S_{\mathcal{C}, s, n, m}$ denote the set of all $m$-samples of concepts $c\in C_{n}$ such that $\textbf{size}(c)\leq s$. For polynomial-time learning algorithm $A$, let $\mathcal{C}^{A}_{s,n, m}\subseteq\mathcal{C}$ denote the $A$-image of $S_{\mathcal{C}, s,n,m}$, i.e., the set of all hypotheses produced by $A$ when $A$ is given an m-sample of a concept $c\in C_{n}$ with $\textbf{size}(c)\leq s$. We also call $\mathcal{C}^{A}_{s,n,m}$ the effective hypothesis space of $A$. Then $A$ is an \textit{Occam algorithm} for $\mathcal{C}$ if $\exists$ polynomial $p(s,n)$ and constant $\alpha$ with $0\leq\alpha<1$ such that $\forall\ s,n,m\geq1$, the VC dimension of $\mathcal{C}^{A}_{s,n,m}$ is bounded by $p(s,n)m^{\alpha}$ \cite{blumer1989learnability}.
\end{definition}

Board and Pitt \cite{board1990necessity}, on the other hand, define their algorithm as follows (some notation has been altered to match the rest of this paper, but the definition itself is equivalent):

\begin{definition}[$\delta$-dependent Occam algorithm]
\label{def:s-delta-dependent-occam}
Let $A$ be a polynomial learning algorithm that, given a sample of a concept $c\in\mathcal{C}$ and parameters $s$ and $\delta$, produces a hypothesis consistent with the sample. For concept class $\mathcal{C}$ and all $s,n,m\geq1$ and $0<\delta<1$, let $S_{\mathcal{C}, s, n, m}$ denote the set of all $m$-samples of concepts $c\in C_{n}$ such that $\textbf{size}(c)\leq s$. For polynomial-time learning algorithm $A$, let $\mathcal{C}^{A}_{s,n, m,\delta}\subseteq\mathcal{C}$ denote the $A$-image of $S_{\mathcal{C}, s,n, m}$, i.e., the set of all hypotheses produced by $A$ when $A$ is given as input $s$, $\delta$, and an m-sample of a concept $c\in C_{n}$ with $\textbf{size}(c)\leq s$. Then $A$ is an \textit{Occam algorithm} for $\mathcal{C}$ if $\exists$ polynomial $p(s,n,\frac{1}{\delta})$ and constant $\alpha$ with $0\leq\alpha < 1$ such that $\forall\ s,n,m\geq1$ and $0<\delta<1$, the VC dimension of $\mathcal{C}^{A}_{s,n,m,\delta}$ is bounded by $p(s,n,\frac{1}{\delta})m^{\alpha}$ \cite{board1990necessity}.
\end{definition}

This is a functional Occam algorithm which takes $s$ and $\delta$ as explicit inputs and has a VC dimension that depends polynomially on $\frac{1}{\delta}$. Each of these factors makes it distinct from the algorithms analysed by Blumer et al. \cite{blumer1989learnability}. We will first show that if we have a functional algorithm dependent on $s$ and $\delta$ as explicit inputs, we can construct a functional algorithm that does not require these inputs explicitly. We will then show that we can construct a $\delta$-independent Occam algorithm from the resulting functional algorithm.

\subsection{Exception lists}
\label{sec:exception lists}

Our analysis will apply only to concept classes which are closed under exception lists.

\begin{definition}\label{def:ex-lists}
    A class $\mathcal{C}$ is closed under exception lists if $\exists$ an algorithm ExList and polynomial $p_{\text{ex}}$ such that $\forall\ n\geq1$, on input of any $c\in C_{n}$ and any finite set $E \subseteq X_{n}$ of cardinality $e$, ExList halts in time bounded by $p_{\text{ex}}(n, s, e)$ and outputs a concept $\text{ExList}(c, E)=c_{E}\in C_{n}$ such that $c_{E}=c\triangle E$. Note that polynomial running time means that $\textbf{size}(c_{E})\leq p_{\text{ex}}(n,s,e)$ (a program cannot output a concept with size larger than its runtime) \cite{board1990necessity}.
\end{definition}

In words, the ExList algorithm will output a concept $c_{E}$ that agrees with $c$ on all but a finite set of points in $X$, with that finite set being given by $E$. Classes that are closed under exception lists include decision lists, decision trees, arbitrary programs, Boolean circuits, Boolean formulas, and Deterministic Finite Automata (DFAs) over any fixed alphabet, among others \cite{board1990necessity}. Many of the above classes happen to be ones where finding a minimal consistent hypothesis is NP-hard, which was the original motivation for developing Occam algorithms \cite{blumer1987occam}.

\section{Equivalence of Occam algorithms}
\label{sec:equivalence}

Board and Pitt's \cite{board1990necessity} construction of an Occam algorithm from a general learning algorithm $L$ begins as follows: ``Run the algorithm $L_{\text{dependent}}$, giving it input parameters $s$, $\epsilon=m^{-\frac{1}{k+1}}$, and $\delta$. Whenever $L_{\text{dependent}}$ calls the oracle for a data point, choose a random point in the sample set $M$ with uniform probability $\frac{1}{m}$ and supply it to $L$ as the example. Let the output of $L$ be denoted $c'$" where we have modified the original notation to be consistent with the rest of this paper and added the label ‘dependent’. By choosing a suitable $k$, we can achieve any value of $\epsilon$ and write it as a polynomial function of $m$.

We know that $\epsilon$ will be given a value that does not require explicit input, as it depends only on $m$. Thus, in practice, the learning algorithm $L_{\text{dependent}}$ in fact only requires $s$, $\delta$, and $M$ as inputs: it is a functional algorithm with $\delta,s$ known. Clearly, this is not one of the learning models considered by \cite{haussler1991equivalence} in Theorem \ref{theorem:equivalence-polynomial}. In contrast, the standard definition of a functional polynomial learning algorithm, Definition \ref{def:functional_poly-learnability}, can be termed a $\delta,s$-unknown functional algorithm. We would like to show an equivalence between the two definitions:

\begin{lem}
\label{lemma:dependent-independent}
If $\mathcal{H}$ is polynomially learnable by a $\delta,s$-known algorithm, it is polynomially learnable by a $\delta,s$-unknown algorithm.
\end{lem}

\begin{prf}
We will construct an $\delta,s$-unknown functional algorithm from a $\delta,s$-known functional algorithm following a method used by Haussler et al. \cite{haussler1991equivalence} to show the equivalence of functional and oracle models of learning.

The $\delta,s$-known functional algorithm defined by Board and Pitt \cite{board1990necessity} is an algorithm that takes $m$ inputs and runs in a polynomially bounded time; thus we need the same for our $\delta,s$-independent functional algorithm. The first requirement is explicit in functional models of learnability, while the latter is explicit in oracle models. Thus, in order to be able to make explicit use of both properties, we will make our algorithm, $L_{\text{independent}}$, a functional algorithm taking input $M$ of length $m$, that has been constructed from an oracle algorithm. We will show that this algorithm then leads to the same result as the one considered by Board and Pitt.

Given that a functional algorithm cannot take $\delta$ or $s$ as input, we need to find a way to bound these values. To do so, we use a construction similar to Haussler et al.'s construction of a functional algorithm from an oracle \cite{haussler1991equivalence}.

If $L_{\text{dependent}}$ has a runtime bounded by polynomial $T_{L}(\epsilon,\delta,n,s)$, we find a polynomial $p$ monotonically increasing in $y$ such that $p(\epsilon,n,y)\geq T_{L}(\epsilon,\frac{1}{y},n,y)$.

We choose a $y$ such that $p(\epsilon,n,y)\leq m$ and $p(\epsilon,n,2y)>m$. We know by definition that $m$ is polynomial in $\frac{1}{\delta}$ and $s$, so this construction means our $y$ is polynomially related to those values. If we cannot do this because $p(\epsilon,n,1)>m$ then we halt with the default hypothesis. Our constructed algorithm $L_{\text{independent}}$ works as follows: we run the algorithm $L_{\text{dependent}}$, giving it input parameters $s_{\text{in}}=y$, $n_{\text{in}}=n$ (which $L$ can determine directly from the form of each example in $M$), $\epsilon_{\text{in}}=\epsilon$, and $\delta_{\text{in}}=\frac{1}{y}$. Whenever $L_{\text{dependent}}$ calls the oracle for a data point, choose a random point in the sample set $M$ with uniform probability $\frac{1}{m}$ and supply it to $L_{\text{dependent}}$ as the example, just as before.

We now claim that $m$ is bounded from below by polynomial $q(\epsilon, \delta, n, s)$. To see this, choose $q$ such that $q(\epsilon,\delta,n,s)=p(\epsilon,n,2(\frac{1}{\delta}+s))$. Then the $y$ we choose will be bounded from below by $\frac{1}{\delta}+s$, so we are running $L_{\text{dependent}}$ with $\delta_{\text{in}}=\frac{1}{y}\leq\delta$ and $s_{\text{in}}=y\geq s$. Thus we have a functional algorithm $L_{\text{independent}}$ whose input and runtime are both bounded polynomially.

This algorithm is equivalent to the one defined by Board and Pitt \cite{board1990necessity}, as it operates in polynomial bounds with $\delta_{\text{in}}=\frac{1}{y}\leq\delta$ and $s_{\text{in}}=y\geq s$. Clearly, if a concept class is learnable by $L_{\text{dependent}}$, then it is learnable by $L_{\text{independent}}$; thus we complete our proof of Lemma \ref{lemma:dependent-independent}.
\end{prf}

Lemma \ref{lemma:dependent-independent} shows that the model of polynomial PAC learnability considered by Board and Pitt is in fact equivalent to the models commonly used in the literature, and is a member of the equivalence class in Theorem \ref{theorem:equivalence-polynomial}. Based on this result, we would now like to construct an Occam algorithm whose output complexity is independent of $\delta$.

\begin{thm}
\label{thm:main}
If $\mathcal{C}$ is polynomially learnable then it is learnable by $\delta$-independent Occam algorithm.
\end{thm}

\begin{prf}
We now repeat Board and Pitt's \cite{board1990necessity} proof of equivalence between general learning and Occam learnability using our newly-constructed $s$- and $\delta$-independent algorithm, $L_{\text{independent}}$, and showing that we still result in a polynomial Occam algorithm, which is now $\delta$-independent.

Define $y$ as in the previous proof. Choose constant $k$ such that we can bound the VC dimension of the hypothesis space, $d(n,T)$ for runtime $T$, with 
\begin{multline}
\label{eq:define-k}
    d(n,T_{L}(\epsilon, \frac{1}{y},n,y))+2\leq d(n, p(\epsilon,n, y))+2\leq\frac{k}{2}\left(\frac{ny^2}{\epsilon}\right)^{k},\\ 
    \forall\ n,y\geq1\text{ and }0<\epsilon<1,
\end{multline}
where $T_{L}$ and $p$ are defined as in the previous proof. Let $\epsilon=m^{-\frac{1}{k+1}}$. Thus, $L_{\text{independent}}$ does not require $\epsilon$, $\delta$, or $s$ as explicit inputs, so it is consistent with the models in Theorem \ref{theorem:equivalence-polynomial}, and if it learns $\mathcal{C}$, we can say that $\mathcal{C}$ is polynomially learnable (as defined in Definition \ref{def:poly-learnable}).

Now, we construct an Occam algorithm $O$ from $L_{\text{independent}}$ as follows:
\begin{enumerate}
    \item Run $L_{\text{independent}}$ as defined in the previous proof, and let its output be denoted $c'$.
    \item Compute the exception list $E = \{x\in M : c(x) \neq c'(x)\}$.
    \item Output $c'_{E} = \text{ExList}(c', E)$.
\end{enumerate}

We define a constant $a_{k}$ such that for any $y\geq a_{k}$, $\log(y)<y^{\frac{1}{k+2}}$. To prove the theorem, we simply need to show that $O$ is an Occam algorithm for $\mathcal{C}$ with corresponding polynomial 
\begin{equation}
\label{eq:occam-characteristic-polynomial}
    p_{O}(n,y)=a_{k}k^{\frac{k+2}{k+1}}(ny^{2})^{\frac{k^{2}+2k}{k+1}}
\end{equation}
and constant exponent
\begin{equation}
\label{eq:occam=characteristic-exponent}
    \alpha=\frac{k^{2}+2k}{k^{2}+2k+1}.
\end{equation}

$L$ runs in polynomial time, as does ExList; thus, $O$ runs in polynomial time. Clearly, any output $c'_{E}$ produced by the algorithm is consistent with $M$. Also, by the definition of $\epsilon$ in Section \ref{sec:concept-classes}, it must be true that $|E|\leq\epsilon|M|$. 

As in Definition \ref{def:s-delta-independent-occam}, let $\mathcal{C}^{O}_{y,n,m}\subseteq \mathcal{C}$ denote the $O$-image of $S_{\mathcal{C},y,n,m}$, i.e., the set of all hypotheses produced by $O$ when $O$ is given an m-sample of a concept $c\in C_{n}$ with $\textbf{size}(c)\leq s$. We would like to find the VC dimension of $\mathcal{C}^{O}_{y,n,m}$.

Let the VC dimension of $\mathcal{C}^{O}_{y,n,m}$ be called $d_{O}$. If $d_{O}\leq1$, it is clearly bounded by $p_{O}(n,y)m^{\alpha}$ and the theorem is proved. Now let us assume $d_{O}\geq2$.

Let $\mathcal{C}^{L}_{y,n,m}$ be the effective hypothesis space of $L_{\text{independent}}$. An algorithm cannot output a hypothesis with $\textbf{size}$ greater than its runtime $T$, so the size of each element of $\mathcal{C}^{L}_{y,n,m}$ is bounded by $T_{L}(\epsilon, \frac{1}{y},n,y)$. Then VC-Dim$(\mathcal{C}^{L}_{y,n,m})\leq d(n,T_{L}(\epsilon,\frac{1}{y},n,y))$.

Making use of Lemma \ref{lemma:exception-list-dimension}, we can write
\begin{equation*}
    \frac{d_{O}}{\log(d_{O})}\leq d(n,T_{L}(\epsilon,\frac{1}{y},n,y))+\epsilon m+2.
\end{equation*}
By our choice of $k$ in \eqref{eq:define-k}, this means that
\begin{equation}\label{eq:d_o/logd_o}
    \frac{d_{O}}{\log(d_{O})}\leq\frac{k}{2}\left(\frac{ny^2}{\epsilon}\right)^{k}+\epsilon m.
\end{equation}

If $d_{O}<a_{k}$ then clearly $d_{O}$ is bounded by $p_{O}(n,y)m^{\alpha}$ and the theorem is proved. 

If $d_{O}\geq a_{k}$ then by choice of $a_{k}$, $\log(d_{O})<(d_{O})^{\frac{1}{k+2}}$. Thus,
\begin{equation*}
    \frac{d_{O}}{\log(d_{O})}>\frac{d_{O}}{(d_{O})^{\frac{1}{k+2}}}=(d_{O})^{\frac{k+1}{k+2}}.
\end{equation*}

Combining with \eqref{eq:d_o/logd_o} we have
\begin{align*}
    (d_{O})^{\frac{k+1}{k+2}}&<\frac{k}{2}\left(\frac{ny^2}{\epsilon}\right)^{k}+\epsilon m\\ 
    &=\frac{k}{2}\left(ny^{2}\right)^{k}\epsilon^{-k} + \epsilon m \\
    &=\frac{k}{2}\left(ny^2\right)^{k}m^{\frac{k}{k+1}}+m^{\frac{k}{k+1}}\\ 
    &\leq k\left(ny^2\right)^{k}m^{\frac{k}{k+1}}.
\end{align*}

Raising each side to $\frac{k+2}{k+1}$, we get
\begin{align*}
    d_{O}& \leq k^{\frac{k+1}{k+2}}(ny^{2})^{\frac{k^{2}+2k}{k+1}}m^{\frac{k^{2}+2k}{k^{2}+2k+1}}\\
    & \leq p_{O}(n,y)m^{\alpha}.
\end{align*}

It should be noted that $y$ and $s$ are polynomially related, so $p_{O}(n,y)=p_{R}(n,s)$ for some polynomial $p_{R}$. Thus, we explicitly satisfy the conditions Definition \ref{def:s-delta-independent-occam}; thus, we have constructed a $\delta$-independent Occam algorithm. This concludes our proof of Theorem \ref{thm:main}.
\end{prf}

\section{Conclusion}
\label{sec:conclusion}

We have now shown that the algorithms used in Board and Pitt's analysis \cite{board1990necessity} are, in fact, equivalent to the algorithmic models used by Blumer et al. \cite{blumer1989learnability}. Thus, the equivalence of learnability and the weak approximability of the minimum-consistent-hypothesis problem is not dependent solely on the choice of algorithmic model, but in fact extends to all algorithmic models considered by Haussler et al. \cite{haussler1991equivalence}.

This analysis focused on strong learning conditions, but based on the equivalence of certain strong and weak learnability criteria, it should naturally extend to those conditions as well. In addition, the same method of proof shows that Board and Pitt's theorem for finite-alphabet length-based $\delta$-dependent Occam algorithms also applies to similarly defined but $\delta$-independent Occam algorithms as used by \cite{blumer1987occam,intro-computational-theory}. Essentially, we have proven an equivalence which had been widely assumed to be true due to the work of Board and Pitt \cite{board1990necessity} but had not, in fact, been shown explicitly to be correct. Thus, we provide \textit{a posteriori} justification of analyses and algorithmic design techniques which made use of this equivalence theorem.


\appendix

\section{Approximate Occam algorithms}

If we relax the condition that Occam algorithms be consistent with the training sample, we can define an ``approximate Occam algorithm" and show that any polynomially PAC learnable concept class is learnable by such an algorithm \cite{board1990necessity}. In fact, it is easy to show that any polynomial PAC learning algorithm can be considered an ``approximate Occam algorithm".

Specifically, we can define approximate Occam algorithms as follows:

\begin{definition}[$\delta$-independent approximate Occam algorithm]
\label{def:s-delta-independent-approximate-occam}
    Let $A$ be a polynomial learning algorithm that, given a sample of a concept $c\in\mathcal{C}$, produces a hypothesis consistent with at least $(1-\epsilon)m$ points in the sample. For concept class $\mathcal{C}$ and all $s,n,m\geq1$, let $S_{\mathcal{C}, s, n, m}$ denote the set of all $m$-samples of concepts $c\in C_{n}$ such that $\textbf{size}(c)\leq s$. For polynomial-time learning algorithm $A$, let $\mathcal{C}^{A}_{s,n, m}\subseteq\mathcal{C}$ denote the $A$-image of $S_{\mathcal{C}, s,n,m}$, i.e., the set of all hypotheses produced by $A$ when $A$ is given an m-sample of a concept $c\in C_{n}$ with $\textbf{size}(c)\leq s$. We also call $\mathcal{C}^{A}_{s,n,m}$ the effective hypothesis space of $A$. Then $A$ is an \textit{Occam algorithm} for $\mathcal{C}$ if $\exists$ polynomial $p(s,n)$ and constant $\alpha$ with $0\leq\alpha<1$ such that $\forall\ s,n,m\geq1$, the VC dimension of $\mathcal{C}^{A}_{s,n,m}$ is bounded by $p(s,n)m^{\alpha}$ \cite{board1990necessity}.
\end{definition}

Then we can show the following lemma:

\begin{lem}
\label{lem:approx-occam}
    If $L$ is a polynomial PAC learning algorithm for $\mathcal{C}$, it is an approximate Occam algorithm for $\mathcal{C}$.
\end{lem}

\begin{prf}
Define $x$ as in the proof of Lemma \ref{lemma:dependent-independent}, and consider a learning algorithm of the type $L_{\text{independent}}$. Then, following the proof of Theorem \ref{thm:main}, we can choose constant $k$ such that we can bound the VC dimension of the hypothesis space, $d(n,T)$ for runtime $T$, with 
\begin{multline}
\label{eq:define-k2}
    d(n,T_{L}(\epsilon, \frac{1}{x},n,x))+2\leq d(n, p(\epsilon,n, x))+2\leq\frac{k}{2}\left(\frac{nx^2}{\epsilon}\right)^{k},\\ 
    \forall\ n,x\geq1\text{ and }0<\epsilon<1,
\end{multline}
where $T_{L}$ and $p$ are defined as in the proof of Lemma \ref{lemma:dependent-independent}. Let $\epsilon=m^{-\frac{1}{k+1}}$. Thus, $L_{\text{independent}}$ does not require $\epsilon$, $\delta$, or $s$ as explicit inputs, so it is consistent with the models in Theorem \ref{theorem:equivalence-polynomial}, and if it learns $\mathcal{C}$, we can say that $\mathcal{C}$ is polynomially learnable (as defined in Definition \ref{def:poly-learnable}).

Let $\mathcal{C}^{L}_{x,n,m}$ be the effective hypothesis space of $L_{\text{independent}}$. An algorithm cannot output a hypothesis with $\textbf{size}$ greater than its runtime $T$, so the size of each element of $\mathcal{C}^{L}_{x,n,m}$ is bounded by $T_{L}(\epsilon, \frac{1}{x},n,x)$. Then 
\begin{align*}
    \text{VC-Dim}(\mathcal{C}^{L}_{x,n,m})&\leq d(n,T_{L}(\epsilon,\frac{1}{x},n,x))\\
    &\leq\frac{k}{2}\left(\frac{nx^2}{\epsilon}\right)^{k}\\
    &=\frac{k}{2}(nx^2)^{k}m^{\frac{k}{k+1}}.
\end{align*}
which is polynomial in $n$ and $x$ and sublinear in $m$. As discussed previously, $x$ is polynomially bounded by $s$, so we have a polynomial in $n$ and $s$ which is multiplied by a sublinear factor in $m$. This completes the proof of Lemma \ref{lem:approx-occam}.
\end{prf}

\section*{Acknowledgements}

I would like to thank Ard Louis for his mentorship and guidance over the course of this work, and for providing feedback on earlier drafts. I also thank the anonymous reviewers for their thoughtful and thorough feedback.

\bibliographystyle{elsarticle-num} 
\bibliography{cas-refs}

\end{document}